\documentclass[letterpaper, 10 pt, conference]{ieeeconf}  

\IEEEoverridecommandlockouts                              
\usepackage[utf8]{inputenc}
\usepackage[T1]{fontenc}
\usepackage{amsmath}
\usepackage{mathrsfs}
\usepackage{amsfonts}
\usepackage{algorithm}
\usepackage{algorithmicx}
\usepackage{algpseudocode}

\usepackage{color}
\usepackage{hyperref}

\usepackage{graphicx}
 
\usepackage{here}
\usepackage{wrapfig}

\title{\LARGE \bf
Whole-body model predictive control with rigid contacts \\ via online switching time optimization 
}

\author{Sotaro Katayama$^{1}$ and Toshiyuki Ohtsuka$^{1}$
\thanks{$^{1}$S. Katayama and T. Ohtsuka are with the Department of System Science, Graduate School of Informatics, Kyoto University, Kyoto, Japan
        {\tt\small katayama@sys.i.kyoto-u.ac.jp}, 
        {\tt\small ohtsuka@i.kyoto-u.ac.jp}}%
}

\begin{document}

\onecolumn
\noindent
© 2022 IEEE. Personal use of this material is permitted. Permission from IEEE must be obtained for all other uses, in any current or future media, including reprinting/republishing this material for advertising or promotional purposes, creating new collective works, for resale or redistribution to servers or lists, or reuse of any copyrighted component of this work in other works.

\hspace{1cm}

\noindent
\textbf{Published article:} \\ 
\noindent
S. Katayama and T. Ohtsuka, ``Whole-body model predictive control with rigid contacts via online switching time optimization,'' 2022 IEEE/RSJ International Conference on Intelligent Robots and Systems (IROS2022), 2022, pp. 8858--8865.

\twocolumn

\maketitle
\thispagestyle{empty}
\pagestyle{empty}

\begin{abstract}

This study presents a whole-body model predictive control (MPC) of robotic systems with rigid contacts, under a given contact sequence using online switching time optimization (STO).
We treat robot dynamics with rigid contacts as a switched system and formulate an optimal control problem of switched systems to implement the MPC.
We utilize an efficient solution algorithm for the MPC problem that optimizes the switching times and trajectory simultaneously.
The present efficient algorithm, unlike inefficient existing methods, enables online optimization as well as switching times.
The proposed MPC with online STO is compared over the conventional MPC with fixed switching times, through numerical simulations of dynamic jumping motions of a quadruped robot.
In the simulation comparison, the proposed MPC successfully controls the dynamic jumping motions in twice as many cases as the conventional MPC, which indicates that the proposed method extends the ability of the whole-body MPC.
We further conduct hardware experiments on the quadrupedal robot Unitree A1 and prove that the proposed method achieves dynamic motions on the real robot.

\end{abstract}

\section{Introduction}

A fundamental difficulty in controlling robotic systems lies in discrete events involved in their dynamics.
Essential tasks of robots such as manipulation and locomotion involve contacts with the environment.
Making and breaking contacts result in switches of their dynamics and impulsive changes in their state.
Therefore, robot systems are inherently modeled as hybrid dynamical systems.
Furthermore, some classes of robots such as quadrupedal robots and humanoid robots are underactuated.
Owing to these difficulties, most analytical model-based controllers focus on only particular classes of problems.
For example, the hybrid zero dynamics controller is a successful approach focusing on bipedal locomotion \cite{bib:hybridZeroDynamicsPlanar, bib:hybridZeroDynamicsCassie}.
However, once we consider more complicated cases: acrobatic jumping, walking on uneven terrain, and multi-contact situations in the real world, such analytical construction of controllers are still difficult.

\begin{figure}
\centering
%

\includegraphics[scale=0.32]{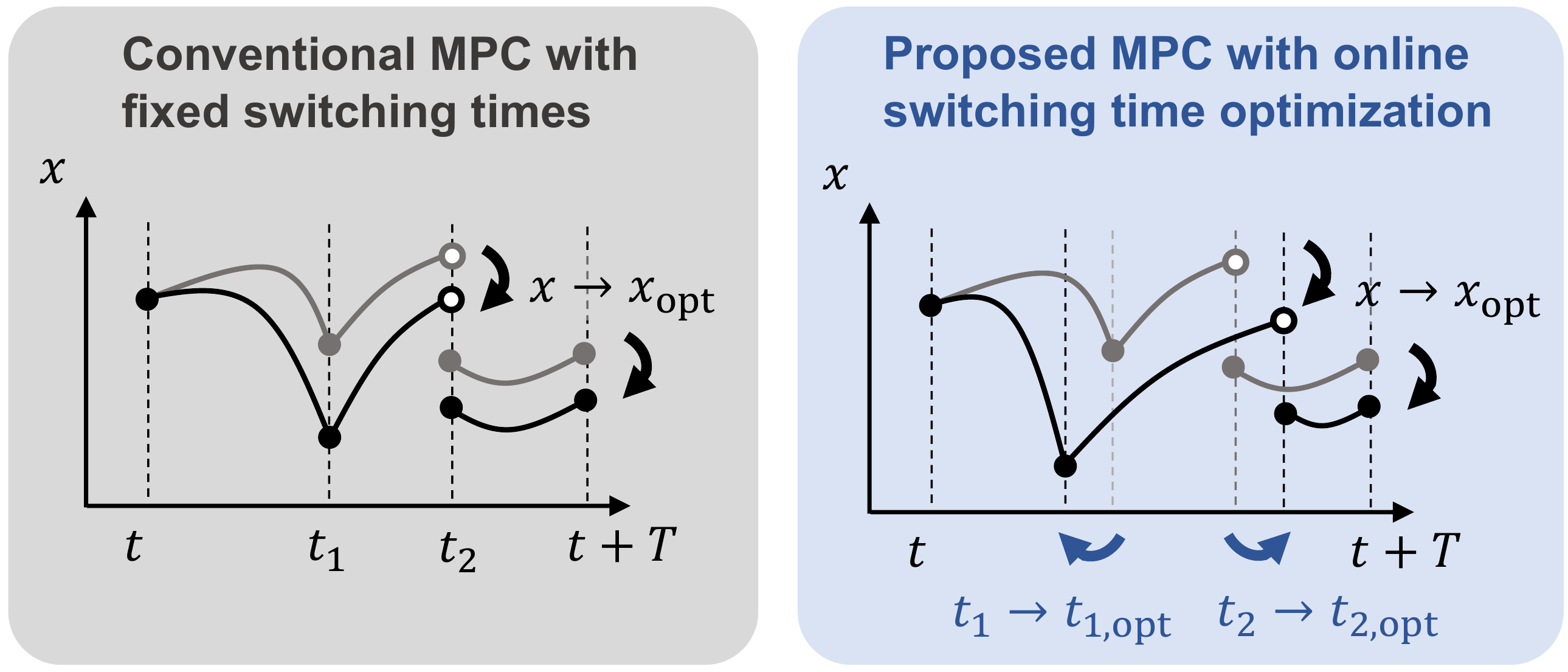}
\caption{
Conceptual diagram of comparison between (left) the conventional model predictive control (MPC) with fixed switching times and (right) the proposed MPC with online switching time optimization. 
The proposed MPC optimizes the switching times $t_1, t_2$ and trajectory $x$ simultaneously, while the conventional MPC solely optimizes the trajectory under the fixed switching times.
}
\label{fig:a1}
\end{figure}

Model predictive control (MPC), which solves trajectory optimization problems online, is expected to be a unified model-based approach to control the robotic systems with contacts; hence, if we can solve the MPC problem for the hybrid systems online, we can realize the robot control systems considering the future discrete event (e.g., interactions with the environment).
However, the hybrid MPC problem is generally formulated as a mixed-integer nonlinear program (MINLP), which is an NP-hard problem owing to its combinatorial nature \cite{bib:MINLP}.
A remarkable approach that can avoid the combinatorial nature is the contact-implicit trajectory optimization (CITO), which is often formulated as mathematical programs with complementarity constraints (MPCC) \cite{bib:MPCC2} or bi-level optimization problems \cite{bib:MPCC1, bib:implicitHardContact}. 
However, MPCC inherently lacks constraint qualifications such as the linear independence constraint qualification (LICQ), and this leads to numerical ill-conditioning.
In addition, numerical alleviations for the LICQ problem often cause undesirable stationary points \cite{bib:MPCCLimits}; thus, it is difficult to even solve MPCC off-line.
Bi-level optimization is also difficult to solve in general, for example, even a linear bi-level optimization problem is NP-hard \cite{bib:bilevelNPhard}.
There are other heuristic approximation methods toward CITO \cite{bib:springDamperCITO, bib:mujoco}, which can be seen as kinds of soft contact models.
However, once the problem involves rigid contacts, these approximation methods can cause non-physical artifacts \cite{bib:validatingSimulators} or their convergence performance can be significantly decreased.

Another tractable and still practical approach is to fix the contact sequence.
In practical situations, a robot has a high-level planner that computes the feasible contact sequence, considering information from perception systems (e.g., safe regions in terrain for foot placements) \cite{bib:PlannningTerrainMapping}.
A low-level motion planner or the MPC controller then optimizes the trajectory, based on the contact sequence.
In this case, the low-level optimization problem is defined, to determine the trajectory and instants of each discrete event (making or breaking the contacts). 
The optimization problem to determine the instants of discrete events is particularly known as the switching time optimization (STO) problem. 
However, it has been difficult to solve the optimal control problem (OCP) involving the STO problem efficiently \cite{bib:twoStage, bib:twoStageConstrained}.
Accordingly, the application of the STO approach remains the offline trajectory optimization of simplified robot models \cite{bib:twoStageApplication1, bib:twoStageApplication2}.
Therefore, recent successful applications of MPC to robotic systems have still been limited to the case with the fixed switching instants \cite{bib:MPCRobot1}. 
However, in practical situations, the behavior of the robot can be different from the predicted one, and the pre-computed switching instants can become infeasible or unreasonable during control, due to model mismatches, disturbances, and dynamic changes in the environment that are not predicted in advance.
In such cases, the control performance deteriorates or the MPC optimizer fails to converge to a feasible solution.

It is worth noting that there are heuristics to adapt the contact timings online \cite{bib:switchingTimeHeuristic1, bib:switchingTimeHeuristic2}.
However, such heuristics can lack versatility; the resultant motions can be conservative and cannot treat multi-objective cost and constraints (for example, \cite{bib:switchingTimeHeuristic1} does not consider the friction cone constraints).
Moreover, the MPC optimizer can still fail to determine a feasible solution with the heuristic timings.

\subsection{Contribution and paper outline}
In this study, we propose a whole-body MPC of robots with rigid contacts using the online STO.
Fig. \ref{fig:a1} illustrates the conceptual diagram of the comparison between the existing MPC framework and proposed MPC.
We utilize an efficient Newton-type algorithm for the OCPs involving STO problems proposed in our previous study \cite{bib:STORiccati}.
The algorithm enables us to efficiently optimize the trajectory and switching times simultaneously.
However, the applicability of the method of \cite{bib:STORiccati} for MPC is not discussed in the previous study, it only demonstrates offline trajectory optimization problems involving the STO problem.
This study implements MPC utilizing the STO algorithm, and proposes several ways to improve the numerical robustness of the MPC, e.g., heuristic regularization and minimum dwell-time constraints.
We conducted numerical simulations on dynamic jumping of a quadrupedal robot and demonstrated that the proposed whole-body MPC with online STO successfully controls the dynamic motions, while conventional MPC with fixed switching times cannot find a feasible solution, thereby failing in the control.
This advantage is observed in various simulation settings, which demonstrates that the proposed method extends the ability of MPC for robots with rigid contacts.
We further conducted hardware experiments of our MPC with online STO on quadruped robot Unitree A1 \cite{bib:a1}, and demonstrated that the proposed MPC with online STO achieved dynamic control of a real robotic system that involves disturbances and model mismatches.

The contributions of this study are then summarized as follows:
\begin{itemize}
    \item To the best of our knowledge, this is the first study that realizes the whole-body MPC of robots with rigid contacts using the online STO. 
    \item We conducted simulation studies on dynamic jumping control of a quadrupedal robot, and demonstrated that the proposed whole-body MPC with the online STO extends the ability of MPC for robots with rigid contacts.
    \item We conducted hardware experiments of our MPC with online STO on quadruped robot Unitree A1. 
\end{itemize}

The remainder of this paper is organized as follows:
Section \ref{section:formulation} models the robot dynamics with rigid contacts as a switched system and formulates its OCP.
Section \ref{section:mpc} describes the MPC with online STO for the OCP described in Section \ref{section:formulation}.
Section \ref{section:simulation} demonstrates the effectiveness of the proposed MPC with online STO over the conventional MPC with fixed switching times, through numerical simulation of the whole-body control of a quadruped robot jumping.
Section \ref{section:experiments} conducts hardware experiments on the quadrupedal robot Unitree A1 and demonstrates that the proposed method achieves dynamic motions on the real robot.
Finally, a brief summary and mention of future studies are presented in Section \ref{section:conclu}.

\section{Optimal Control Problem Formulation}\label{section:formulation}
\subsection{Rigid body systems with rigid contacts}
First, we model a rigid body system with rigid contacts (a robot having contacts with the environment) as a switched system, and formulate its OCP, as \cite{bib:DDPContact, bib:crocoddyl, bib:twoStageApplication2}.
To formulate an OCP of the switched systems without the combinatorial nature or complementarity constraints, we assume that the contact sequence (the sequence of the active contacts) is given, for example, it is provided by a higher-level planner.
Let $q, v, a \in \mathbb{R}^n$, $\lambda \in \mathbb{R}^{n_{\lambda}}$, and $u \in \mathbb{R}^{m}$ be the configuration, generalized velocity, acceleration, stack of the contact forces, and joint torques, respectively. 
Note that the presented formulation can consider the configuration space including $SE(3)$ to model the floating base as \cite{bib:liftedContactDynamics}, while this study assumes the Euclidian configuration space for notational simplicity.
Also note that the contact dimension $n_{\lambda}$ alters depending on combinations of the active contacts.
The equation of motion of the rigid-body system is then expressed as:
\begin{equation}\label{eq:equationOfMotion}
    M(q) a + h(q, v) - J^{\rm T} (q) \lambda = S^{\rm T} u,
\end{equation}
where $M (q) \in \mathbb{R}^{n \times n}$ denotes the inertia matrix, $h (q, v) \allowbreak \in \mathbb{R}^{n}$ encompasses the Coriolis, centrifugal, and gravitational terms, $J (q) \in \mathbb{R}^{n_{\lambda} \times n}$ denotes the stack of the contact Jacobians, and $S \in \mathbb{R}^{m \times n}$ denotes the selection matrix. 
We assume that $J(q)$ is non-singular throughout this paper.
Because the contact sequence is provided, we can treat the contact constraints as a bilateral constraint of the form 
\begin{equation}\label{eq:p}
    {\bf p} (q) = 0,
\end{equation}
where ${\bf p} (q) \in \mathbb{R}^{n_{\lambda}}$ is the stack of the positions (and includes rotations for surface contacts) of the active contact frames.
Furthermore, instead of considering (\ref{eq:p}) over a time interval, we consider the constraint on the frame acceleration the over the time interval \cite{bib:baumgarte}:
\begin{subequations}\label{subeq:baum}
\begin{equation}\label{eq:a}
    {\bf a} (q, v, a) := \ddot{\bf p} + 2 \alpha \dot{\bf p} + \beta ^2 {\bf p} = J (q) a + {\bf b} (q, v),
\end{equation}
where $\alpha$ and $\beta$ are weight parameters, and we define
\begin{equation}\label{eq:b}
  {\bf b} (q, v) := \dot{J} (q, v) v + 2 \alpha J (q) v + \beta^2 {\bf p} (q).
\end{equation}
\end{subequations}
Then the original position constraint (\ref{eq:p}) is satisfied over the time interval provided that (\ref{eq:p}) and the equality constraint on the contact velocity, 
\begin{equation}\label{eq:v}
    \dot{\bf p} (q, v) = J(q) v = 0,
\end{equation}
are satisfied at some points in the interval \cite{bib:baumgarteParameters}.
Equation (\ref{subeq:baum}) is reduced to twice the time derivative of (\ref{eq:p}) with $\alpha = \beta = 0$; however, the constraint violation of the original position constraint (\ref{eq:p}) can be accumulated because of the numerical computation.
$\alpha = \beta > 0$ is typically chosen to stabilize the violation of the original constraint (\ref{eq:p}).
By combining (\ref{eq:equationOfMotion}) and (\ref{subeq:baum}), we obtain the contact-consistent forward dynamics:
\begin{equation}\label{eq:contactDynamics}
    \begin{bmatrix}
        a \\ 
        - \lambda
    \end{bmatrix}
    = 
    \begin{bmatrix}
        M (q) & J^{\rm T} (q) \\ 
        J (q) & O
    \end{bmatrix}^{-1}
    \begin{bmatrix}
        S ^{\rm T} u - h (q, v) \\
        - {\bf b} (q, v)
    \end{bmatrix}.
\end{equation}
We define the state vector and state equation as 
\begin{equation}\label{eq:stateEquation}
    x := 
    \begin{bmatrix}
        q \\
        v
    \end{bmatrix},
    \;\;\;
    \dot{x} = f (x, u) 
    := 
     \begin{bmatrix}
        v \\
        a (x, u)
    \end{bmatrix} ,
\end{equation}
respectively, where $a (x, u) \in \mathbb{R}^n$ is a function of $x$ and $u$ as defined in (\ref{eq:contactDynamics}).

We would like to emphasize that the state equation (\ref{eq:stateEquation}) switches depending on the combination of active contacts, which we also refer to ``contact mode'' in the following. 
That is, the rigid body system with contacts is a switched system.
We express the state equations (\ref{eq:stateEquation}) for each contact mode as
\begin{equation*}\label{eq:stateEquationSwitch}
    \dot{x} = f_{k} (x, u), 
\end{equation*}
where $k$ denotes the index of the contact mode.
We refer to $k$ as the index of an active subsystem of the switched system in the following.

When the bodies of the system and environment collide, the generalized velocity of the system alters according to the equation of Newton's law of impact:
\begin{equation}\label{eq:equationOfImpulse}
    M (q) \delta v - J ^{\rm T} (q) \Lambda = 0,
\end{equation}
where $\delta v \in \mathbb{R}^n$ denotes the impulsive change in the generalized velocity and $\Lambda \in \mathbb{R}^{n_{\lambda}}$ denotes the stack of the impact forces.
The evolution of the state between the impact is expressed as
\begin{equation}\label{eq:qvEvolutionImpulse}
\begin{bmatrix}
    q^+ \\
    v^+
\end{bmatrix}
= \begin{bmatrix}
    q^- \\ 
    v^- + \delta v
\end{bmatrix},
\end{equation}
where $q^-, v^- \in \mathbb{R}^n$ denote the configuration and velocity immediately before the impulse and $q^+, v^+ \in \mathbb{R}^n$ do immediately after the impulse, respectively. 
The contact position constraint (\ref{eq:p}) for the frames with the impact is also imposed at the impulse instant, which corresponds to switching or guard conditions of the hybrid systems \cite{bib:hybridZeroDynamicsPlanar, bib:hybridZeroDynamicsCassie}.
We assume a completely inelastic collision, in that the contact velocity constraints (\ref{eq:v}) holds for the velocity immediately after the impulse $v^+ = v^- + \delta v$, i.e., 
\begin{equation}\label{eq:vplus}
    {\bf v} (q^-, v^-, \delta v) := \dot{\bf p} (q^-, v^- + \delta v) = J (q^-) (v^- + \delta v) = 0.
\end{equation}
By combining (\ref{eq:equationOfImpulse}) and (\ref{eq:vplus}), we obtain the impulse dynamics:
\begin{equation}\label{eq:impulseDynamics}
    \begin{bmatrix}
        \delta v \\ 
        - \Lambda
    \end{bmatrix}
    = 
    \begin{bmatrix}
        M (q^-) & J^{\rm T} (q^-) \\ 
        J (q^-) & O
    \end{bmatrix}^{-1}
    \begin{bmatrix}
        0 \\
        - J (q^-) v^-
    \end{bmatrix}.
\end{equation}
With these settings, the state jump equation is expressed as
\begin{equation}\label{eq:jumpEquation}
    x ^+
    :=
    \begin{bmatrix}
        q ^+ \\ 
        v ^+ 
    \end{bmatrix}
    = \psi (x^-)
    = \begin{bmatrix}
        q^- \\
        v ^- + \delta v ^- (x^-)
    \end{bmatrix},
\end{equation}
where $\delta v^- (x^-) \in \mathbb{R}^n$ is a function of $x^-$, as defined in (\ref{eq:impulseDynamics}).
Note that the impulse forces occur only for bodies that have impacts with the environment.
Therefore, (\ref{eq:p}) and (\ref{eq:jumpEquation}) change depending on the contact modes immediately before and after the collision ($k-1$ and $k$), which are expressed as 
\begin{equation*}\label{eq:switchingConditionSwitch}
    {\rm \bf p}_{k-1, k} (x^-) = 0
\end{equation*}
and
\begin{equation*}\label{eq:jumpEquationSwitch}
    x ^+ = \psi_{k-1, k} (x^-), 
\end{equation*}
respectively.

\subsection{Inequality constraints}
The robot system involves several physical limitations, such as joint position, velocity, and torque limits.
Furthermore, contact force must lie inside the friction cone; otherwise, the resultant solution can be physically-infeasible (for example, the solution allows negative-direction normal contact forces).
We thus introduce polyhedral-approximated friction cone constraint for each contact force $[\lambda_x \;\, \lambda_y \;\, \lambda_z]$ that is expressed in the contact surface frame as:
\begin{equation}\label{eq:frictionCone}
    \begin{bmatrix}
        \lambda_x + \frac{\mu}{\sqrt{2}} \lambda_z\\
      - \lambda_x + \frac{\mu}{\sqrt{2}} \lambda_z\\
      \lambda_y + \frac{\mu}{\sqrt{2}} \lambda_z \\
        - \lambda_y + \frac{\mu}{\sqrt{2}} \lambda_z \\
        \lambda_z
    \end{bmatrix} \geq 0, 
\end{equation}
where $\mu > 0$ is the friction coefficient.
Note that the friction cone constraint (\ref{eq:frictionCone}) is solely imposed for each active contact.
Therefore, the collection of the inequality constraints imposed at a time instant also switches depending on the contact situations.
We summarize the collection of the inequality constraints for each contact mode (that is, for each active subsystem $k$) as
\begin{equation*}\label{eq:inequalityConstraintsSwitch}
    g_{k} (x, u) \leq 0.
\end{equation*}

\subsection{Optimal control problem of switched systems}
We herein summarize the modeling of the rigid body system as a switched system and formulate an OCP for the system.
Without loss of generality, let $\mathcal{K} = \left\{ 1, ..., K+1 \right\}$ ($K > 0$) be the given indices of active subsystems over the horizon of the OCP, i.e., $\mathcal{K}$ is the given contact sequence.
The continuous-time OCP is provided as follows.
We consider a switched system comprising $K+1$ subsystems
\begin{subequations}
\begin{equation}\label{eq:stateEquationMode}
    \dot{x} (t) = f_{k} (x(t), u(t)), \;\; t \in [t_{k-1}, t_{k}), \;\; k \in \mathcal{K}, 
\end{equation}
with state jumps and switching conditions
\begin{equation}\label{eq:stateJumpEquationMode}
    x (t_k) = \psi_{k-1, k} (x(t_{k}-)), \; p_{k-1, k} (x(t_{k}-)) = 0, \;\; j \in \mathcal{K}_j, 
\end{equation}
and constraints 
\begin{equation}\label{eq:constraintsMode}
    g_{k} (x(t), u(t)) \leq 0, \;\; t \in [t_{k-1}, t_{k}], \;\; k \in \mathcal{K} ,
\end{equation}
\end{subequations}
where $\mathcal{K}_j \subset \mathcal{K}$ denotes the set of contact modes that involve impacts between the system and environment at the switch from $k \in \mathcal{K}_j$ to $k+1$. 
Subsequently, $\mathcal{K} - \mathcal{K}_j$ denotes the indices that do not involve state jumps, i.e., the switch from $k \in \mathcal{K} - \mathcal{K}_j$ to $k+1$ only involve breaking the contacts.
Note that we also refer to the index $k$ as ``phase'' in this study.
$t_0$ and $t_{K+1}$ denote the fixed initial and terminal times of the horizon and $t_k$, $k \in \left\{ 1, ..., K \right\}$ denotes the switching instant from phase $k$ to phase $k+1$.
For a given initial state $x(t_0) \in \mathbb{R}^{n_x}$, we consider the OCP of the switched system of the form of 
\begin{subequations}\label{subeq:continuousTimeOCP}
\begin{equation}\label{eq:costFunction}
    \min_{u(\cdot), t_{1}, ..., t_{K}} {J} = \; V_f (x(t_{K+1})) + \sum_{k=1}^{K+1} \int_{t_{k-1}}^{t_{k}} l_k (x(\tau), u(\tau)) d \tau
\end{equation}
\begin{align}
    {\rm s.t.} \;\;\; & {\rm (\ref{eq:stateEquationMode})-(\ref{eq:constraintsMode})}, \; \notag \\ 
    & t_{k-1} + \underline{\Delta}_{k} \leq t_{k}, \;\; k \in \mathcal{K} \label{eq:minimumDwellTimes}
\end{align}
\end{subequations}
where $V_f (\cdot)$ and $l_k (\cdot)$ are user-defined terminal cost and stage cost for phase $k$, respectively. 
$\underline{\Delta}_{k} \geq 0, \; k \in \mathcal{K}$ is the minimum dwell-time, i.e., the last constraints in (\ref{eq:minimumDwellTimes}) are the minimum dwell-time constraints.

\section{Model Predictive Control with Online Switching Time Optimization}\label{section:mpc}
\subsection{Direct multiple shooting method with mesh-refinement}

\begin{figure}[tb]
\centering
\includegraphics[scale=0.37]{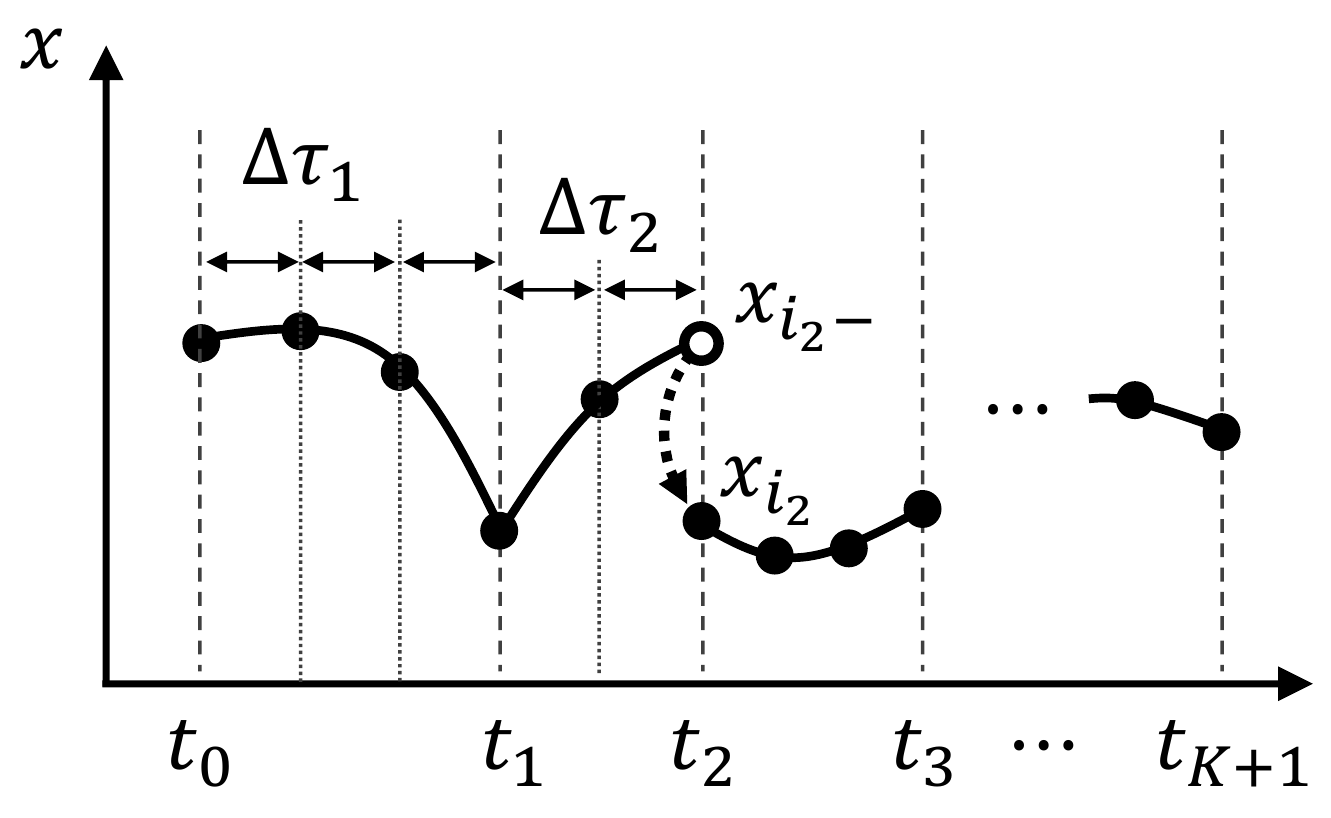}
\caption{Conceptual diagram of the proposed discretization method of the continuous-time OCP of a switched system. Solid lines illustrate the continuous transition of the state. The switch from phase 2 to phase 3 involves a state jump (i.e., $2 \in \mathcal{K}_j$), which is illustrated as a dotted curve arrow. A white circle indicates a new grid representing the state immediately before the state jump.}
\label{fig:discretization}
\end{figure}

The OCP (\ref{subeq:continuousTimeOCP}) includes the switching instants as the optimization variables, as well as the state and control trajectory.
The first key point for efficient and robust numerical computation is the discretization method of the continuous-time OCP.
We discretize the continuous-time OCP using the direct multiple shooting (DMS) method \cite{bib:DMS} to equalize the all-time steps in a phase. A conceptual diagram of the discretization method is illustrated in Fig. \ref{fig:discretization}.
We index the grid corresponding to the $k$-th switch as $i_k$ and that immediately before the switch as $i_k-$ if the switch involves the state jump.
We then introduce $N$ grid points over the horizon, discretized state $X := \left\{x_0, ..., x_N, x_{i_k-} , ..., x_{i_j-} \right\}$, discretized control input $U := \left\{ u_0, ..., u_{N-1} \right\}$, and switching instants $T:= \left\{t_1, ..., t_K \right\}$.
The resultant nonlinear program (NLP) is given as 
\begin{subequations}\label{subeq:NLP}
\begin{equation}\label{eq:NLP:cost}
    \min_{X, U, T} J = V_f (x_N) + \sum_{k \in \mathcal{K}} \sum_{i \in \mathcal{I}_k} l_k (x_i, u_i) \Delta \tau_k
\end{equation}
\begin{align}
    {\rm s.t.} \;\;\; & x_0 - \bar{x} = 0 , \label{eq:NLP:initialStateConstraint} \\
    & x_i + f_k (x_i, u_i) \Delta \tau_k - x_{i+1} = 0, \;\; i \in \mathcal{I}_k, \; k \in \mathcal{K} , \label{eq:NLP:stateEquation} \\ 
    & x_{i_k} = f_{j} (x_{i_k -}), \; p_{j} (x_{i_k -}) = 0, \;\; k \in \mathcal{K}_j, \label{eq:NLP:stateJumpEquation} \\
    & g_k (x_i, u_i) \leq 0, \;\; i \in \mathcal{I}_k, \; k \in \mathcal{K}, \label{eq:NLP:inequalityConstraints} \\
    & t_{k-1} + \underline{\Delta}_k - t_{k} \leq 0, \;\; k \in \left\{ 1, ..., K \right\} \label{eq:NLP:STOConstraints},
\end{align}
where $\Delta \tau_k$ is the time step at phase $k$, defined as
\begin{equation}
    \Delta \tau_k := \frac{t_{k} - t_{k-1}}{N_k}, \;\; k \in \left\{ 1, ..., K \right\}.
\end{equation}
\end{subequations}
An advantage over the two-stage methods \cite{bib:twoStage}, which have been applied to robot systems in \cite{bib:twoStageApplication1, bib:twoStageApplication2}, of the NLP formulation (\ref{subeq:NLP}) is that the local convergence of the Newton-type method for the NLP (\ref{subeq:NLP}) is guaranteed.
After solving the NLP (\ref{subeq:NLP}), each discretization step-size $\Delta \tau_k$ can change before solving it because the switching times are optimized.
Therefore, after the NLP, we check the step-size $\Delta \tau_k$ for the all-phase $k \in \mathcal{K}$. 
If it is too large, we perform mesh-refinement to increase the accuracy of the solution in terms of the continuous-time counterpart, i.e., we add grids on the phase $k$ where $\Delta \tau_k$ is large. 

\subsection{Riccati recursion to compute Newton step}
We solve the NLP (\ref{subeq:NLP}) using primal-dual interior point method \cite{bib:nocedal} with Gauss-Newton Hessian approximation.
The Newton-step computation (solution of a linear system) is then reduced to that of the unconstrained OCP by eliminating the Newton-step of the slack and dual variables of the inequality constraints. 
Subsequently, the Riccati recursion algorithm for the OCP of switched systems proposed in \cite{bib:STORiccati} is adopted for the Newton-step computation.
The method has the following three characteristics: 
\begin{itemize}
    \item Its computational time is linear-time complexity with regards to the number of the discretization grids $N$. The per Newton-step computation is approximately the same computational cost as the conventional Riccati recursion algorithm for Newton-type methods of unconstrained OCPs, which is efficient for large scale systems, therefore, it is often adopted for complicated robotic applications \cite{bib:twoStageApplication1, bib:twoStageApplication2, bib:DDPContact, bib:crocoddyl}.
    \item It only requires the positive semi-definiteness of the reduced Hessian matrix that is obtained by multiplying the null-space matrix of the Jacobian of constraints to the Hessian matrix. Therefore, the Riccati recursion algorithm can treat the NLP (\ref{subeq:NLP}) whose Hessian matrix is inherently indefinite.
    Conversely, some general QP solvers require the positive definiteness of the Hessian matrix and cannot treat the NLP (\ref{subeq:NLP}). 
    \item It improves the numerical stability by modifying the reduced Hessian matrix positive definite without additional computational burden. 
\end{itemize}
We treat the pure-state constraints (\ref{eq:p}) efficiently within the Riccati recursion algorithm using a constraint-transformation of \cite{bib:stateConstrainedRiccati}.
Furthermore, we lift the contact-consistent forward dynamics (\ref{eq:contactDynamics}) and impulse dynamics (\ref{eq:impulseDynamics}) to relax the high nonlinearity of the NLP and improve convergence property, without increasing the additional computational burden \cite{bib:liftedContactDynamics}.

\subsection{Heuristic regularization to improve convergence property}
Even with the Hessian modification of the Riccati recursion algorithm \cite{bib:STORiccati}, the Newton-step computation can be ill-conditioned owing to the non-convex nature of the NLP.
We then heuristically introduce a large penalty on the update of the switching times when the iterate is not close to a local minimum.
That is, we add a large constant to the diagonal element of the Hessian matrix corresponding twice to the partial derivatives with respect to a switching time.

\subsection{Minimum dwell-time constraints}

\begin{figure}[tp]
\centering
\includegraphics[scale=0.23]{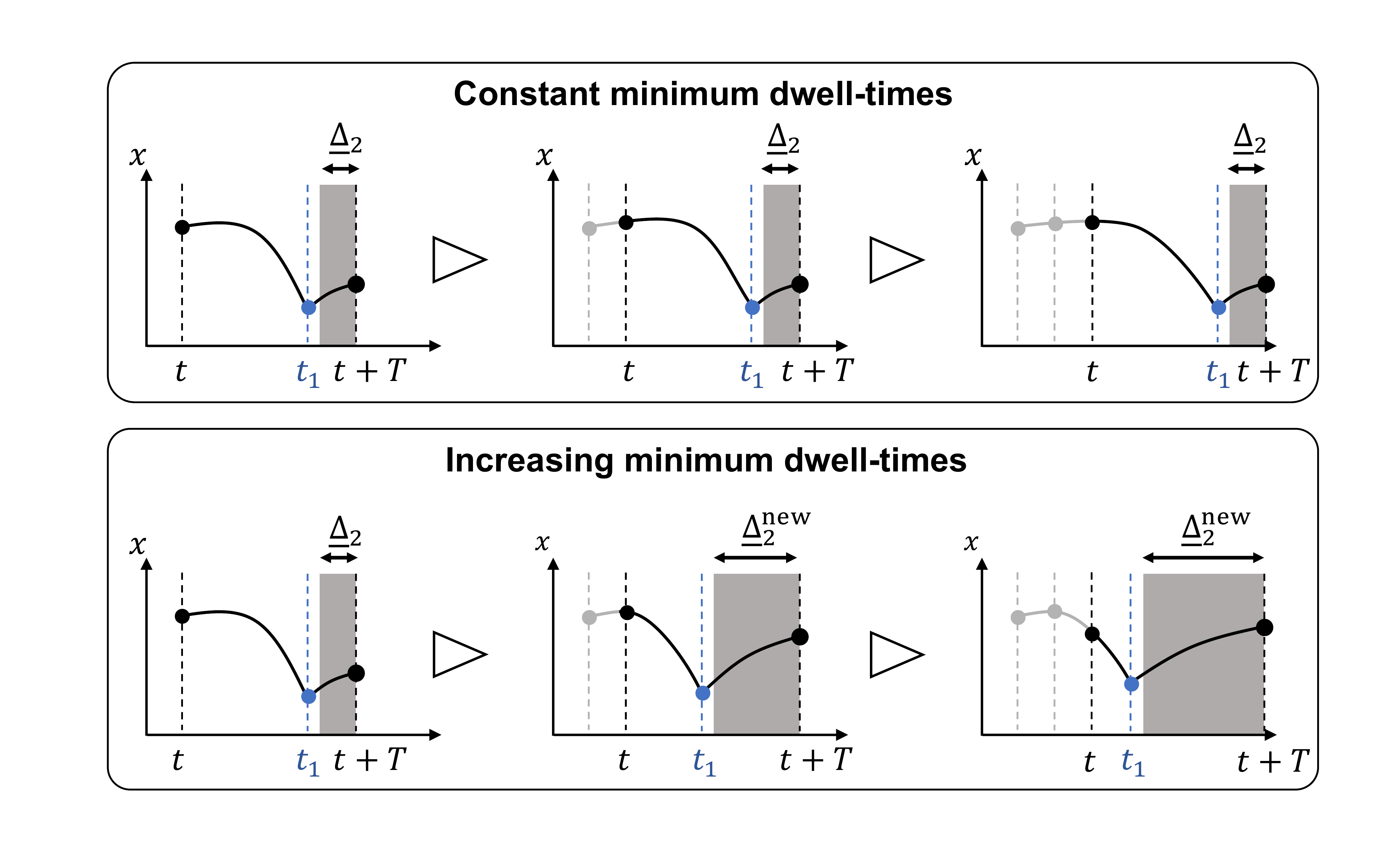}
\caption{
Effects of increasing the minimum dwell-time constraints. The upper figure illustrates a situation where the switching time $t_1$ is kept at the end of the horizon owing to the fixed minimum dwell-time. The lower figure illustrates a solution for this problem by increasing the minimum dwell-time $\underline{\Delta}_{K+1}$.
$t$ denotes the sampling instant, i.e., the initial time of the horizon, and $T$ denotes the length of the horizon.
}
\label{fig:minimumDwellTime}
\end{figure}

The minimum dwell-time constraints (\ref{eq:minimumDwellTimes}) play an important role in practice.
We observe that the convergence speed increases as the minimum dwell-time $\underline{\Delta}_k$ increases, because it reduces search ranges of $t_k$.
However, too large $\underline{\Delta}_k$ can reduce the optimality and ill condition the problem.
Therefore, we have to carefully choose $\underline{\Delta}_k$.

Moreover, in MPC implementation, the optimized switching times can be procrastinated depending on the problem settings (e.g., the given contact sequence and cost function). 
That is, with certain problem settings, the cost of the switching is high and the switching times always lie near the terminal of the horizon. 
As a result, predicted discrete events (i.e., making or breaking the contacts) never happen on the real system.
This issue is illustrated in the upper figure of Fig. \ref{fig:minimumDwellTime}.
To prevent this phenomenon in MPC, at each sampling time, we gradually increase the minimum-dwell time of the last phase $\underline{\Delta}_{K+1}$, which is illustrated in the lower figure of Fig. \ref{fig:minimumDwellTime}. 
For example, at each sampling time, we increase $\underline{\Delta}_{K+1}$ as 
\begin{equation}\label{eq:minDwelltimeUpdate}
    \underline{\Delta}_{K+1} ^{\rm new} \leftarrow \underline{\Delta}_{K+1} + ({\rm sampling} \;\, {\rm period}).
\end{equation}

\subsection{Software implementation}
We implement the proposed MPC algorithm as an open-source software \texttt{robotoc}, an efficient optimal control solver for robotic systems \cite{bib:robotocWeb}, which is written in C++, uses Eigen for linear algebra, OpenMP for stage-wise parallel computation of the cost, constraints, and their derivatives of the NLP, and Pinocchio \cite{bib:pinocchio} for the rigid body dynamics and its derivatives computations.

\section{Simulation Study: Comparison to Conventional MPC with Fixed Contact Timings}\label{section:simulation}
\subsection{Experimental settings and MPC design}
We demonstrated the effectiveness of the proposed whole-body MPC with online STO over the conventional whole-body MPC with the fixed switching instants through numerical simulations.
We refer to the former as ``MPC-STO'' and the latter just as ``MPC'' in the following.
We considered the $0.6$ m jumping of quadrupedal robot Unitree A1, a torque-controlled quadruped robot whose degree of freedom (DOF) is 12. 
We designed the cost functions $V_f (\cdot)$ and $l_k (\cdot)$ as simple quadratic weights on the deviation of the configuration from the reference standing pose, generalized velocity, and acceleration. Note that the acceleration is defined as the function of $x$ and $u$ as (\ref{eq:contactDynamics}). 
That is, we did not impose any costs on the pre-defined jump trajectory but just specified landing contact locations by the contact constraints (\ref{eq:p})--(\ref{eq:b}).
The jumping motions were induced mainly by these contact constraints.
We imposed inequality constraints comprising the joint position, velocity, torque limits, and the polyhedral-approximated friction cone constraints (\ref{eq:frictionCone}). 
We set the MPC settings as follows: the horizon length was $0.8$ s, the number of the discretization grids $N$ was 20, and the number of the Newton-type iterations per sampling time was 2.
We fixed the barrier parameter of the primal-dual interior point method as $1.0 \times 10^{-3}$ throughout the simulation, which is a common and practical strategy of suboptimal MPC \cite{bib:MPCBoyd}.
We provided the same initial guess of the solution including the switching instants to the two MPC controllers.
We investigated the control results for various initial guesses of the switching instants, i.e., lift-off and touch-down times.
This investigation corresponds to the situation where pre-computed optimal switching time is not appropriate owing to disturbances.
We ran the simulations on physical simulator PyBullet \cite{bib:pybullet} and observed the state directly from the simulator.
The MPC-STO and MPC at a 400 Hz sampling rate (2.5 ms sampling period).
In MPC-STO, we updated the minimum-dwell time constraints according to (\ref{eq:minDwelltimeUpdate}). 

\begin{figure*}[htp]
\centering
\includegraphics[scale=0.25]{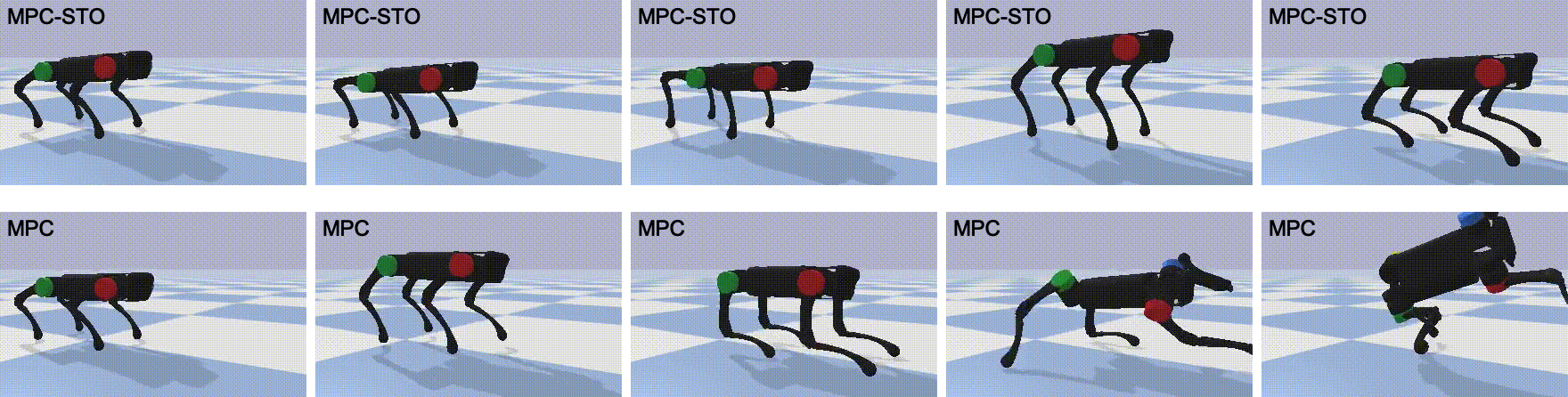}
\caption{Snapshots of the 0.6 m jumping control simulations of quadruped robot Unitree A1 by whole-body MPC-STO (proposed) and whole-body MPC (conventional), in which the initial guesses of lift-off and touch-down times are given by 0.2 s and 0.5 s, respectively. 
The upper successful simulation (labeled as MPC-STO) is with the proposed whole-body MPC controller with online STO. The lower failing simulation (labeled as MPC) is with the conventional whole-body MPC controller with fixed switching times.
Pictures with the same column are taken from the same simulation instant.
A supplemental video is found at \url{https://youtu.be/SureDVDFbfM}.
}
\label{fig:snapshots}
\end{figure*}

\begin{figure}[htp]
\centering
\includegraphics[scale=0.75]{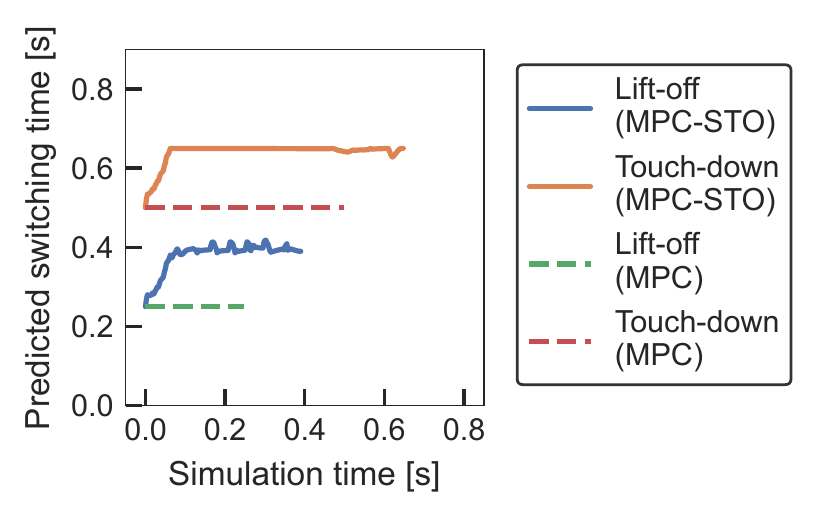}
\caption{Predicted switching times at each simulation time of whole-body MPC-STO (proposed) and whole-body MPC (conventional) for whole-body jumping control of quadruped robot Unitree A1 in which the initial guesses of lift-off and touch-down times are given by 0.2 s and 0.5 s, respectively.}
\label{fig:STO}
\end{figure}

\begin{figure}[htp]
\centering
\includegraphics[scale=0.75]{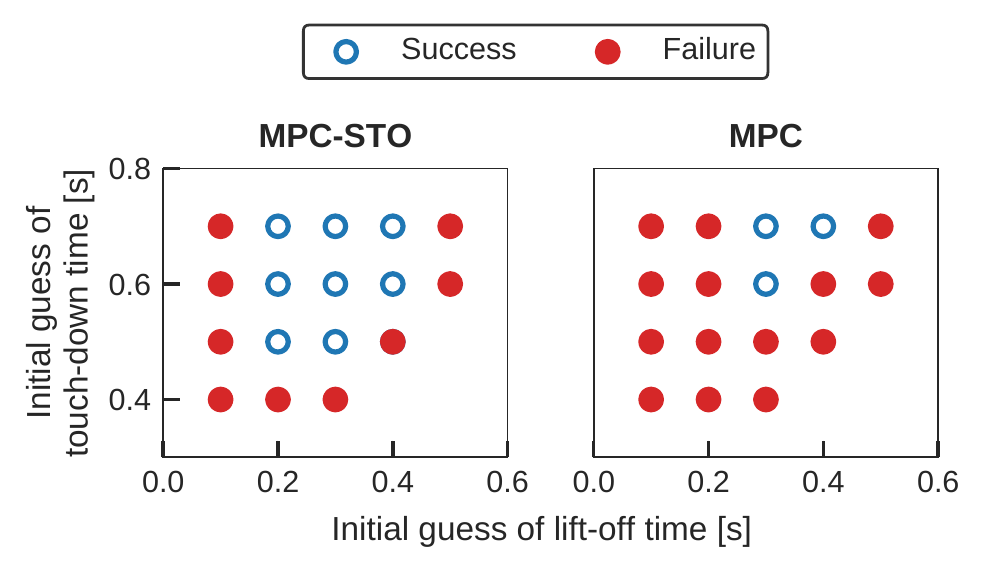}
\caption{Simulation results (success or failure) of the whole-body MPC-STO (proposed) and whole-body MPC (conventional) for whole-body jumping control of quadruped robot Unitree A1 for various initial guesses of the lift-off and touch-down times.}
\label{fig:success}
\end{figure}

\subsection{Results}
Fig. \ref{fig:snapshots} illustrates the snapshots of the simulation results in which the initial guesses of the lift-off and touch-down times are given by 0.2 s and 0.5 s, respectively. 
The upper five pictures are taken from the proposed whole-body MPC controller labeled as ``MPC-STO''.
The lower five pictures are from the proposed whole-body MPC controller labeled as ``MPC''.
As illustrated in Fig. \ref{fig:snapshots}, the conventional MPC failed in whole-body jumping control of the quadruped, whereas our MPC-STO succeeded.
Fig. \ref{fig:STO} illustrates the predicted switching times of both the proposed MPC-STO and conventional STO at each simulation time.
As illustrated in Fig. \ref{fig:STO}, the proposed method optimized the switching times immediately, which led to success in the control of the dynamic jumping motion.

Fig. \ref{fig:success} illustrates the control results of the proposed MPC-STO and conventional MPC controllers for various initial guesses of the switching times.
The proposed MPC-STO succeeded more than twice as many cases as the conventional MPC, which demonstrates that the proposed method extends the ability of whole-body MPC.
This means that the proposed MPC-STO can achieve various control even though there are large disturbances that make the pre-computed switching instants meaningless.
The average computational time of the control updates at each sampling time of both MPC and MPC-STO was approximately 1.3 ms, i.e, each Newton-type iteration took 0.65 ms, on octa-core CPU Intel Core i9-11900H @2.50 GHz with six threads in the parallel computation of the DMS.
In the failure cases of MPC-STO, the solver could not converge to the optimal solution owing to the limited number of Newton-type iterations per sampling time (2 Newton-type iterations).
If we can reduce the computational time per Newton-type iteration, the controller can take a greater number of the iterations, and we then expect that MPC-STO can succeed even with worse initial guesses.
Therefore, improving the computational speed is still an important future study.

\begin{figure}[t]
\centering
\includegraphics[scale=0.21]{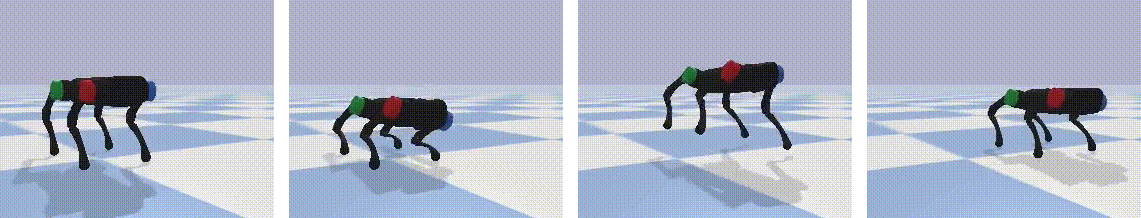}

\vspace{2.0mm}

\includegraphics[scale=0.21]{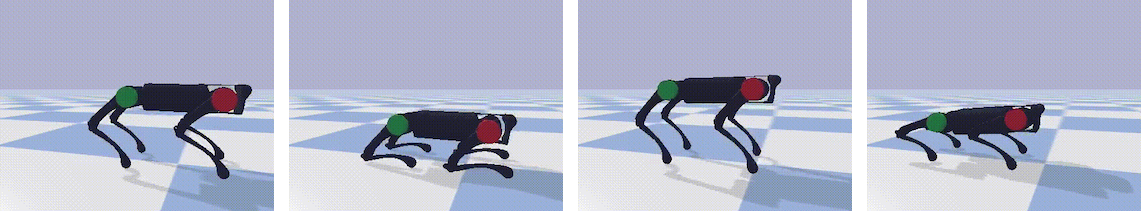}

\vspace{2.0mm}

\includegraphics[scale=0.21]{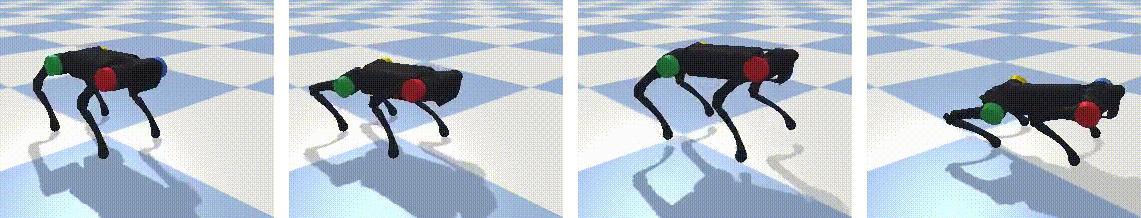}
\caption{Snapshots of dynamic 0.4 m lateral jumping (upper), 0.3 m jumping to back (middle), and 30-degree yaw-rotational jumping (lower) control of quadrupedal robot Unitree A1 by whole-body MPC-STO.}
\label{fig:lateralAndRotationalJumping}
\end{figure}

Note that the present MPC-STO can easily be applied to other kinds of dynamic motions.
To show this, we further conducted simulations of several jumping motions.
We used the almost same MPC design as the above experiments: we only changed the landing contact positions and reference pose of the configuration in the cost function.
Fig. \ref{fig:lateralAndRotationalJumping} shows the snapshots of dynamic lateral jumping, jumping to back, and rotational jumping by the proposed whole-body MPC controller.
Our MPC algorithm with online STO successfully controlled these motions only with the simple cost function without efforts to construct complicated cost functions specialized to each motion or heuristics to determine the switching times.

\section{Hardware Experiments on Quadrupedal Robot Unitree A1}\label{section:experiments}
\subsection{Experimental settings and controller design}
\begin{figure}
\centering
\includegraphics[scale=0.37]{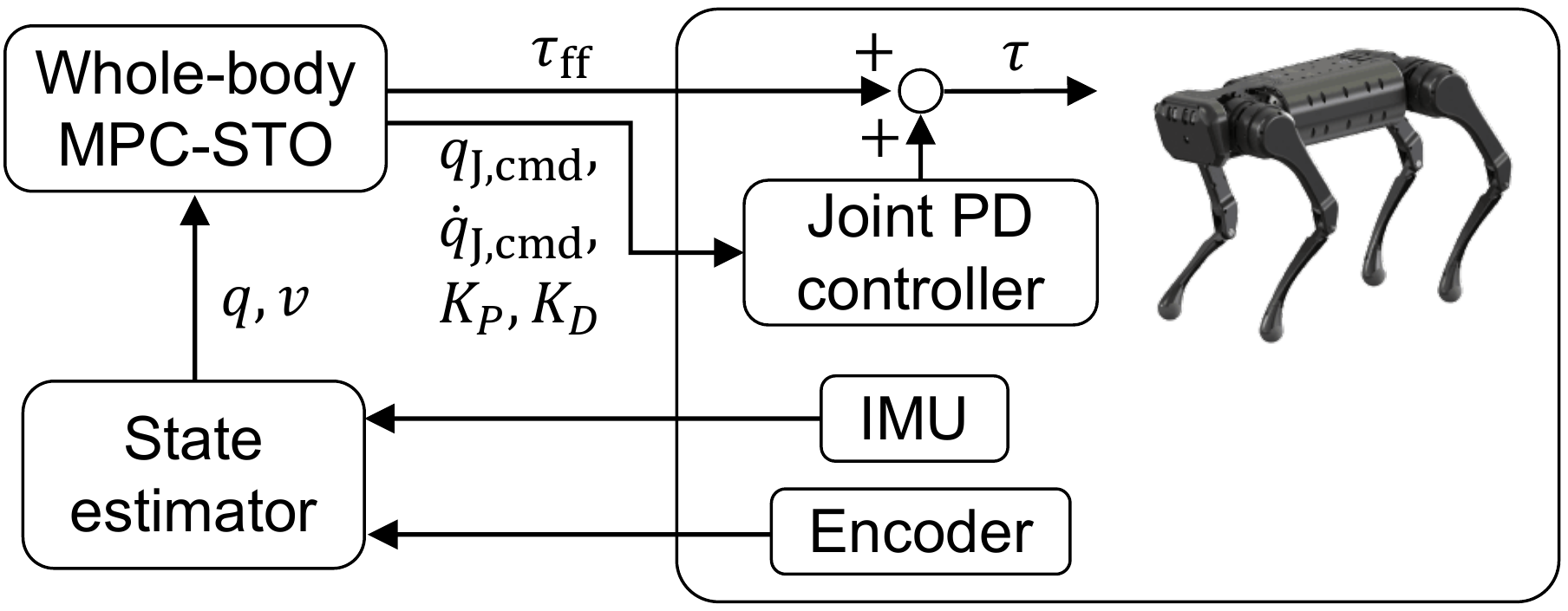}
\caption{Block diagram of the whole-body MPC-STO control framework of quadruped robot Unitree A1.}
\label{fig:controller}
\end{figure}

We further validated the proposed MPC on the real hardware of the quadrupedal robot Unitree A1. 
We applied the proposed MPC for successive dynamic jumps of the quadruped robot.
The overall control framework is illustrated in Fig. \ref{fig:controller}.
We implemented a state estimator to estimate the state of the floating base using encoders and an inertial measurement unit (IMU).
Specifically, we estimate the contact state from joint measurements \cite{bib:contacEstimation} and then estimate the state of the floating base by fusing IMU measurements and the contact leg kinematics as in \cite{bib:InEKFLegged}.
We also implemented a cascaded controller using the whole-body MPC-STO and joint PD controllers to compensate model-mismatches such as joint frictions and estimation errors.
That is, we send the desired joint angle and joint velocity computed by the MPC-STO as well as the feedforward joint torques ($q_{\rm J, cmd}$, $\dot{q}_{\rm J, cmd}$, and $\tau_{\rm ff}$ in Fig. \ref{fig:controller}, respectively) to the build-in joint PD controller of the robot.
We also utilize the state-feedback gain obtained by the Riccati recursion algorithm of the MPC-STO as the joint PD gains ($K_P$ and $K_D$ in Fig. \ref{fig:controller}) because it can be seen as the optimal state-feedback gain of a linear-quadratic approximation of the NLP (\ref{subeq:NLP}).

The MPC-STO and state estimator are run at 400 Hz on off-board laptop Ubuntu, with the PREEMPT RT kernel on octa-core CPU Intel Core i9-11900H @2.50 GHz with six threads in the parallel computing for the DMS in the MPC.
The joint PD controller is run on the onboard PC of Unitree A1.

\subsection{Results}

\begin{figure*}[htp]
\centering
\includegraphics[scale=0.52]{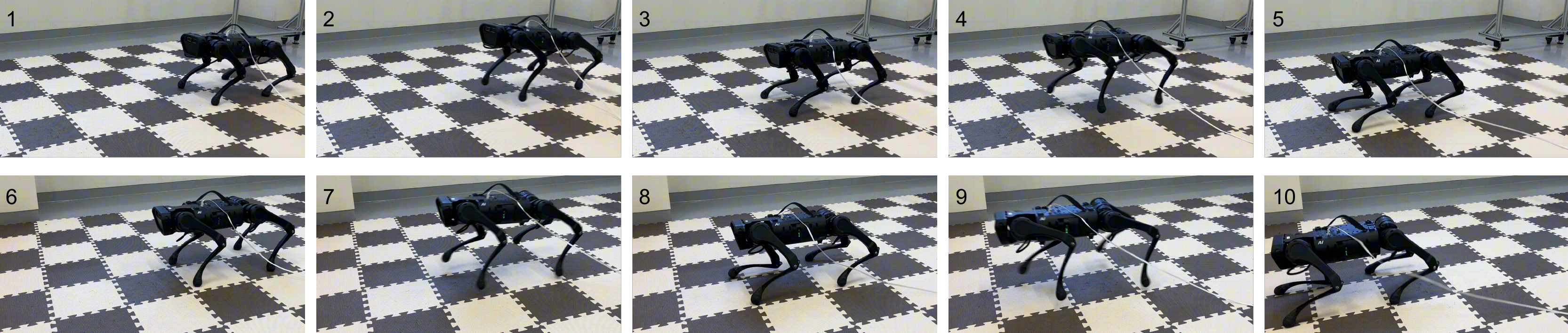}
\caption{Snapshots of successive four-times dynamic jumping of quadruped robot Unitree A1 by the proposed whole-body MPC-STO. 
A supplemental video is found at \url{https://youtu.be/SureDVDFbfM}.
}
\label{fig:snapshots_hw}
\end{figure*}

\begin{figure}[htp]
\centering
\includegraphics[scale=0.625]{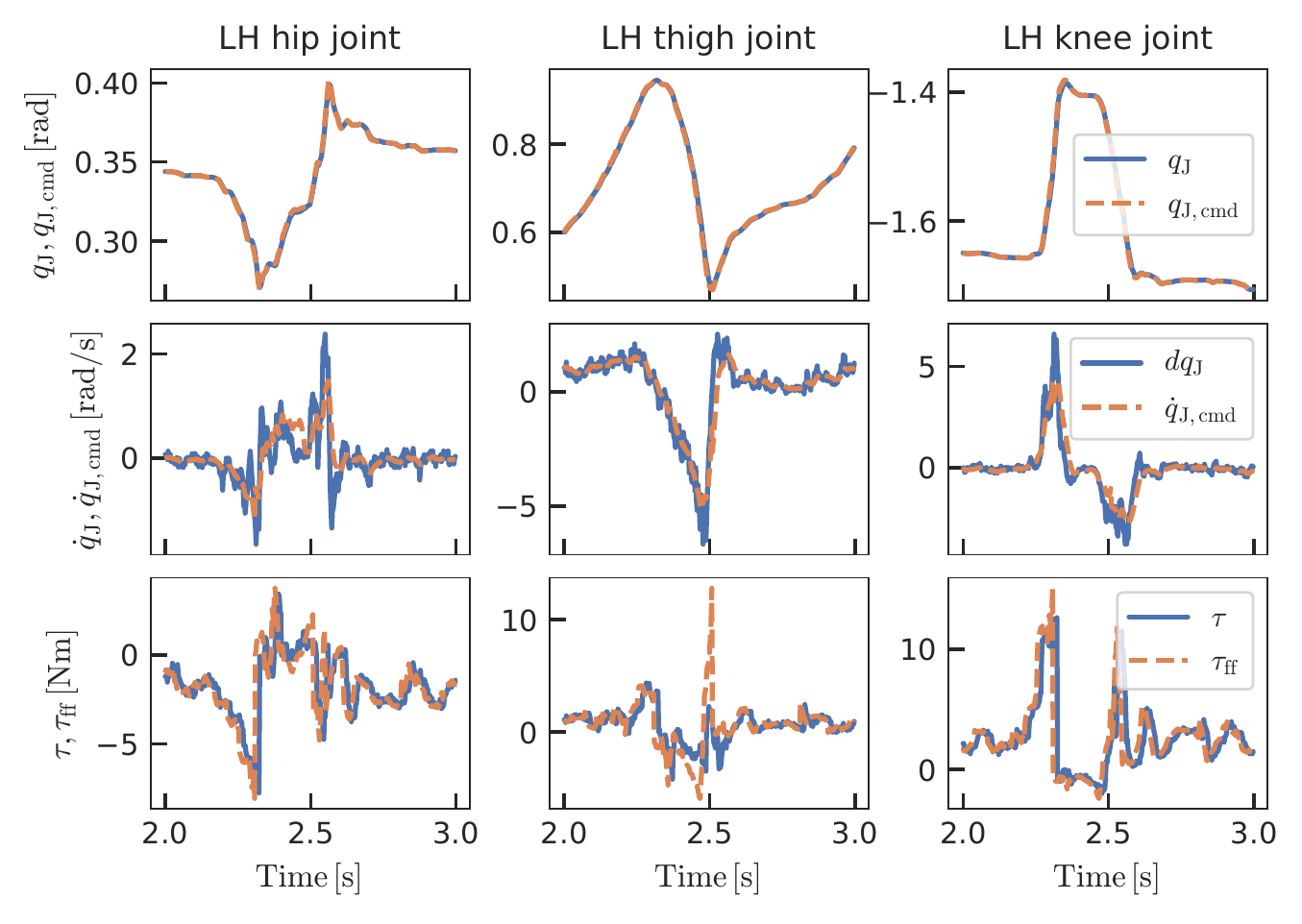}
\caption{A part of the time histories of measurements and commands of the joints in LH leg in the dynamic jumping control.
}
\label{fig:timeHistories}
\end{figure}

Fig. \ref{fig:snapshots_hw} illustrates the successive four-times dynamic jumping via the proposed whole-body MPC controller.
As illustrated in the figure, the proposed method successfully realized the dynamic motions of the quadruped robot.
Fig. \ref{fig:timeHistories} shows a part of the time-histories of measurements and commands of three joints in the LH-leg.
It showed that each motor could not exactly track the commanded position, velocity, and torques provided by the proposed MPC due to the un-modeled factors such as joint frictions, i.e., disturbances. 
In addition, there could be state-estimation errors and modeling errors in the robot models.
Nevertheless, the proposed controller achieved the control thanks to the fast MPC feedback and cascaded joint PD controller. 


\section{Conclusions}\label{section:conclu}
In this study, we presented a whole-body MPC of robotic systems with rigid contacts under a given contact sequence using the online STO.
We treated the robot dynamics with rigid contacts as a switched system and formulated an OCP of switched systems to implement the MPC.
We utilized the efficient solution algorithm for the MPC involving the STO problem that optimizes the switching times and trajectory simultaneously.
The present algorithm is efficient to enable online optimization, unlike the existing methods because of their inefficiency.
We demonstrated the effectiveness of the proposed MPC with online STO over the conventional MPC with fixed switching times, through numerical simulation on dynamic jumping motions of a quadruped robot.
Accordingly, the proposed MPC succeeded in the control more than twice as many cases as the conventional MPC.
We further conducted hardware experiments on the quadrupedal robot Unitree A1 and demonstrated that the proposed method achieves dynamic motions on the real robot.

A possible limitation of our method is that it requires the pre-defined contact sequence and contact locations.
The methodology for their determination and synthesis, with the proposed STO strategy are included in future studies.







\section*{Acknowledgment}
This work was partly supported by JST SPRING, Grant Number JPMJSP2110, and JSPS KAKENHI, Grant Numbers JP22J11441 and JP22H01510.


\bibliographystyle{IEEEtran}
\bibliography{IEEEabrv, ieee}

\begin{thebibliography}{10}
\providecommand{\url}[1]{#1}
\csname url@samestyle\endcsname
\providecommand{\newblock}{\relax}
\providecommand{\bibinfo}[2]{#2}
\providecommand{\BIBentrySTDinterwordspacing}{\spaceskip=0pt\relax}
\providecommand{\BIBentryALTinterwordstretchfactor}{4}
\providecommand{\BIBentryALTinterwordspacing}{\spaceskip=\fontdimen2\font plus
\BIBentryALTinterwordstretchfactor\fontdimen3\font minus
  \fontdimen4\font\relax}
\providecommand{\BIBforeignlanguage}[2]{{%
\expandafter\ifx\csname l@#1\endcsname\relax
\typeout{** WARNING: IEEEtran.bst: No hyphenation pattern has been}%
\typeout{** loaded for the language `#1'. Using the pattern for}%
\typeout{** the default language instead.}%
\else
\language=\csname l@#1\endcsname
\fi
#2}}
\providecommand{\BIBdecl}{\relax}
\BIBdecl

\bibitem{bib:hybridZeroDynamicsPlanar}
E.~Westervelt, J.~Grizzle, and D.~Koditschek, ``Hybrid zero dynamics of planar
  biped walkers,'' \emph{IEEE Transactions on Automatic Control}, vol.~48,
  no.~1, pp. 42--56, 2003.

\bibitem{bib:hybridZeroDynamicsCassie}
J.~Reher and A.~D. Ames, ``Inverse dynamics control of compliant hybrid zero
  dynamic walking,'' in \emph{2021 IEEE International Conference on Robotics
  and Automation (ICRA)}, 2021, pp. 2040--2047.

\bibitem{bib:MINLP}
P.~Belotti, C.~Kirches, S.~Leyffer, J.~Linderoth, J.~Luedtke, and A.~Mahajan,
  ``Mixed-integer nonlinear optimization,'' \emph{Acta Numerica}, vol.~22, p.
  1–131, 2013.

\bibitem{bib:MPCC2}
M.~Posa, C.~Cantu, and R.~Tedrake, ``A direct method for trajectory
  optimization of rigid bodies through contact,'' \emph{The International
  Journal of Robotics Research}, vol.~33, no.~1, pp. 69--81, 2014.

\bibitem{bib:MPCC1}
K.~Yunt, ``An augmented {L}agrangian based shooting method for the optimal
  trajectory generation of switching {L}agrangian systems,'' \emph{Dynamics of
  Continuous, Discrete and Impulsive Systems Series B: Applications and
  Algorithms}, vol.~18, no.~5, pp. 615--645, 2011.

\bibitem{bib:implicitHardContact}
J.~{Carius}, R.~{Ranftl}, V.~{Koltun}, and M.~{Hutter}, ``Trajectory
  optimization with implicit hard contacts,'' \emph{IEEE Robotics and
  Automation Letters}, vol.~3, no.~4, pp. 3316--3323, 2018.

\bibitem{bib:MPCCLimits}
A.~Nurkanovi{\'c}, S.~Albrecht, and M.~Diehl, ``Limits of {MPCC} formulations
  in direct optimal control with nonsmooth differential equations,'' in
  \emph{2020 European Control Conference (ECC)}, 2020, pp. 2015--2020.

\bibitem{bib:bilevelNPhard}
P.~Hansen, B.~Jaumard, and G.~Savard, ``New branch-and-bound rules for linear
  bilevel programming,'' \emph{SIAM Journal on Scientific and Statistical
  Computing}, vol.~13, no.~5, p. 1194–1217, 1992.

\bibitem{bib:springDamperCITO}
M.~Neunert, M.~Stäuble, M.~Giftthaler, C.~D. Bellicoso, J.~Carius, C.~Gehring,
  M.~Hutter, and J.~Buchli, ``Whole-body nonlinear model predictive control
  through contacts for quadrupeds,'' \emph{IEEE Robotics and Automation
  Letters}, vol.~3, no.~3, pp. 1458--1465, 2018.

\bibitem{bib:mujoco}
E.~Todorov, ``Convex and analytically-invertible dynamics with contacts and
  constraints: Theory and implementation in {MuJoCo},'' in \emph{2014 IEEE
  International Conference on Robotics and Automation (ICRA)}, 2014, pp.
  6054--6061.

\bibitem{bib:validatingSimulators}
B.~Acosta, W.~Yang, and M.~Posa, ``Validating robotics simulators on real-world
  impacts,'' \emph{IEEE Robotics and Automation Letters}, vol.~7, no.~3, pp.
  6471--6478, 2022.

\bibitem{bib:PlannningTerrainMapping}
C.~Mastalli, I.~Havoutis, M.~Focchi, D.~G. Caldwell, and C.~Semini, ``Motion
  planning for quadrupedal locomotion: {C}oupled planning, terrain mapping, and
  whole-body control,'' \emph{IEEE Transactions on Robotics}, vol.~36, no.~6,
  pp. 1635--1648, 2020.

\bibitem{bib:twoStage}
X.~Xu and P.~J. {Antsaklis}, ``Optimal control of switched systems based on
  parameterization of the switching instants,'' \emph{IEEE Transactions on
  Automatic Control}, vol.~49, no.~1, pp. 2--16, 2004.

\bibitem{bib:twoStageConstrained}
J.~Fu and C.~Zhang, ``Optimal control of path-constrained switched systems with
  guaranteed feasibility,'' \emph{IEEE Transactions on Automatic Control},
  2021.

\bibitem{bib:twoStageApplication1}
F.~Farshidian, M.~Kamgarpour, D.~Pardo, and J.~Buchli, ``Sequential linear
  quadratic optimal control for nonlinear switched systems,''
  \emph{IFAC-PapersOnLine}, vol.~50, no.~1, pp. 1463 -- 1469, 2017, 20th IFAC
  World Congress.

\bibitem{bib:twoStageApplication2}
H.~{Li} and P.~M. {Wensing}, ``Hybrid systems differential dynamic programming
  for whole-body motion planning of legged robots,'' \emph{IEEE Robotics and
  Automation Letters}, vol.~5, no.~4, pp. 5448--5455, 2020.

\bibitem{bib:MPCRobot1}
J.-P. Sleiman, F.~Farshidian, M.~V. Minniti, and M.~Hutter, ``A unified {MPC}
  framework for whole-body dynamic locomotion and manipulation,'' \emph{IEEE
  Robotics and Automation Letters}, vol.~6, no.~3, pp. 4688--4695, 2021.

\bibitem{bib:switchingTimeHeuristic1}
S.~Caron and Q.-C. Pham, ``When to make a step? tackling the timing problem in
  multi-contact locomotion by {TOPP}-{MPC},'' in \emph{2017 IEEE-RAS 17th
  International Conference on Humanoid Robotics (Humanoids)}, 2017, pp.
  522--528.

\bibitem{bib:switchingTimeHeuristic2}
F.~M. Smaldone, N.~Scianca, L.~Lanari, and G.~Oriolo, ``Feasibility-driven step
  timing adaptation for robust mpc-based gait generation in humanoids,''
  \emph{IEEE Robotics and Automation Letters}, vol.~6, no.~2, pp. 1582--1589,
  2021.

\bibitem{bib:STORiccati}
\BIBentryALTinterwordspacing
S.~Katayama and T.~Ohtsuka, ``Structure-exploiting {N}ewton-type method for
  optimal control of switched systems,'' 2021. [Online]. Available:
  \url{arXiv:2112.07232}
\BIBentrySTDinterwordspacing

\bibitem{bib:a1}
``A1 website,'' \url{https://www.unitree.com/products/a1/}.

\bibitem{bib:DDPContact}
R.~Budhiraja, J.~Carpentier, C.~Mastalli, and N.~Mansard, ``Differential
  dynamic programming for multi-phase rigid contact dynamics,'' in
  \emph{IEEE-RAS International Conference on Humanoid Robots (ICHR)}, 2018.

\bibitem{bib:crocoddyl}
C.~Mastalli, R.~Budhiraja, W.~Merkt, G.~Saurel, B.~Hammoud, M.~Naveau,
  J.~Carpentier, L.~Righetti, S.~Vijayakumar, and N.~Mansard, ``Crocoddyl: An
  efficient and versatile framework for multi-contact optimal control,'' in
  \emph{IEEE International Conference on Robotics and Automation (ICRA)}, 2020,
  pp. 2536--2542.

\bibitem{bib:liftedContactDynamics}
S.~Katayama and T.~Ohtsuka, ``Lifted contact dynamics for efficient optimal
  control of rigid body systems with contacts,'' in \emph{2022 IEEE/RSJ
  International Conference on Intelligent Robots and Systems (IROS)
  (accepted)}, 2022.

\bibitem{bib:baumgarte}
J.~Baumgarte, ``Stabilization of constraints and integrals of motion in
  dynamical systems,'' \emph{Computer Methods in Applied Mechanics and
  Engineering}, vol.~1, no.~1, pp. 1--16, 1972.

\bibitem{bib:baumgarteParameters}
P.~Flores, M.~Machado, E.~Seabra, and M.~Silva, ``A parametric study on the
  {Baumgarte} stabilization method for forward dynamics of constrained
  multibody systems,'' \emph{Journal of Computational and Nonlinear Dynamics},
  vol.~6, p. 011019, 2011.

\bibitem{bib:DMS}
H.~Bock and K.~Plitt, ``A multiple shooting algorithm for direct solution of
  optimal control problems,'' in \emph{9th IFAC World Congress}, 1984, pp.
  1603--1608.

\bibitem{bib:nocedal}
J.~Nocedal and S.~J. Wright, \emph{Numerical Optimization}, 2nd~ed.\hskip 1em
  plus 0.5em minus 0.4em\relax Springer, 2006.

\bibitem{bib:stateConstrainedRiccati}
S.~Katayama and T.~Ohtsuka, ``Efficient {Riccati} recursion for optimal control
  problems with pure-state equality constraints,'' in \emph{{2022 American
  Control Conference (ACC)}}, 2022, pp. 3579--3586.

\bibitem{bib:robotocWeb}
\BIBentryALTinterwordspacing
S.~Katayama, ``robotoc,'' 2020--2022. [Online]. Available:
  \url{https://github.com/mayataka/robotoc}
\BIBentrySTDinterwordspacing

\bibitem{bib:pinocchio}
J.~Carpentier, G.~Saurel, G.~Buondonno, J.~Mirabel, F.~Lamiraux, O.~Stasse, and
  N.~Mansard, ``The {P}inocchio {C}++ library -- {A} fast and flexible
  implementation of rigid body dynamics algorithms and their analytical
  derivatives,'' in \emph{International Symposium on System Integration (SII)},
  2019, pp. 614 -- 619.

\bibitem{bib:MPCBoyd}
Y.~Wang and S.~Boyd, ``Fast model predictive control using online
  optimization,'' \emph{IEEE Transactions on Control Systems Technology},
  vol.~18, no.~2, pp. 267--278, 2010.

\bibitem{bib:pybullet}
E.~Coumans and Y.~Bai, ``Pybullet, a python module for physics simulation for
  games, robotics and machine learning,'' \url{http://pybullet.org},
  2016--2022.

\bibitem{bib:contacEstimation}
M.~Camurri, M.~Fallon, S.~Bazeille, A.~Radulescu, V.~Barasuol, D.~G. Caldwell,
  and C.~Semini, ``Probabilistic contact estimation and impact detection for
  state estimation of quadruped robots,'' \emph{IEEE Robotics and Automation
  Letters}, vol.~2, no.~2, pp. 1023--1030, 2017.

\bibitem{bib:InEKFLegged}
R.~Hartley, M.~Ghaffari, R.~M. Eustice, and J.~W. Grizzle, ``Contact-aided
  invariant extended {K}alman filtering for robot state estimation,'' \emph{The
  International Journal of Robotics Research}, vol.~39, no.~4, pp. 402--430,
  2020.

\end{thebibliography}


\end{document}